\documentclass[10pt,twocolumn,letterpaper]{article}

\usepackage{iccv}
\usepackage{times}
\usepackage{epsfig}
\usepackage{graphicx}
\usepackage{amsmath}
\usepackage{amssymb}
\usepackage{enumitem}
\usepackage{subfig}

\usepackage[breaklinks=true,bookmarks=false]{hyperref}

\iccvfinalcopy 

\def\iccvPaperID{6337} 

\ificcvfinal\fi
\begin{document}
\newcommand{\be}{\begin{equation}}
\newcommand{\ee}{\end{equation}}

\title{Toroidal AutoEncoder}


\author{Maciej Mikulski \qquad Jaroslaw Duda \\
Jagiellonian University\\
Krakow, Poland\\
{\tt\small maciej.mikulski.jr@gmail.com \qquad jaroslaw.duda@uj.edu.pl}
}
\maketitle

\begin{abstract}
Enforcing distributions of latent variables in neural networks is an active subject. It is vital in all kinds of generative models, where we want to be able to interpolate between points in the latent space, or sample from it. Modern generative AutoEncoders (AE) like WAE, SWAE, CWAE add a regularizer to the standard (deterministic) AE, which allows to enforce Gaussian distribution in the latent space. Enforcing different distributions, especially topologically nontrivial, might bring some new interesting possibilities, but this subject seems unexplored so far.

This article proposes a new approach to enforce uniform distribution on d-dimensional torus. We introduce a circular spring loss, which enforces minibatch points to be equally spaced and satisfy cyclic boundary conditions.

As example of application we propose multiple-path morphing. Minimal distance geodesic between two points in uniform distribution on latent space of angles becomes a line, however, torus topology allows us to choose such lines in alternative ways, going through different edges of $[-\pi,\pi]^d$.

Further applications to explore can be for example trying to learn real-life topologically nontrivial spaces of features, like rotations to automatically recognize 2D rotation of an object in picture by training on relative angles, or even 3D rotations by additionally using spherical features - this way morphing should be close to object rotation.
\end{abstract}



\section{Introduction}
	
Autoencoders are a class of models that learn to recreate its input on its output passing it through a bottleneck. In computer vision they are used to encode images into an useful latent space. Instead of directly working on pixels, it allows to work on latent representation of the image. It is also possible to give a meaning to some of latent variables, e.g. the color of the hair or sex of the person on the image. Ability to change those features, even perform an arithmetics on them (\cite{ari1,ari2}), and then decode the image back to the space of pixels is a very powerful tool. One can also just pick a random feature vector in the latent space and generate a realistically looking image. A very interesting task is to interpolate (morph) between two images encoded in the feature space and see how the output pictures changes.

The main research interest of this article is answering a question  how to enforce AE to learn a chosen distribution in the latent space. Why is it a problem? Training a vanilla (i.e. without any regularization of the latent space) AE will lead to very irregular distribution: some dimensions will explore just very narrow areas, some will be broad, many will be correlated. We can imagine the learnt distribution as a manifold that we could walk around and change the features. But for vanilla AE we would quickly fall from it or at least break our neck. This is where all *AE methods arrive:

The original Variational AutoEncoder~\cite{vae} have used a nondeterministic encoder, leading to additional blurring. More recent approaches, like WAE~\cite{wae}, SWAE~\cite{swae}, CWAE~\cite{cwae}, GAE~\cite{gae}, use standard deterministic autoencoder, adding to reconstruction loss a regularizer evaluating distance between probability distribution on latent space of mini-batch, and the chosen prior distribution - usually Gaussian.

\begin{figure}[h!]
	\centering
		\includegraphics[width=0.45\textwidth]{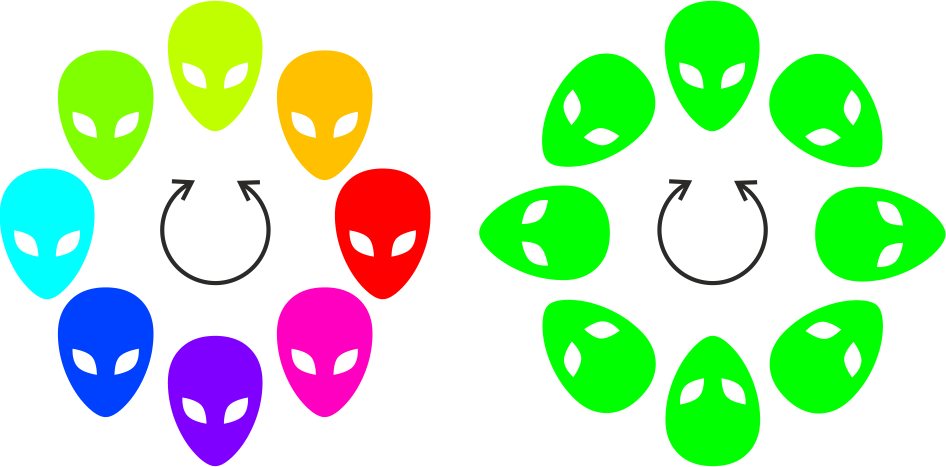}
	\caption{There exists more than one way to morph or rotate between two images. Such multiple circular features topologically form a torus.}
	\label{fig:faces_circ}
\end{figure}

Methods mentioned above usually aim at forcing AE to use Gaussian distribution on the latent space. This is usually a successful approach, because many real world features (like size of eyes) represent Gaussian distribution in the population. But there are some features where Gaussian distribution is not suitable, especially if having nontrivial topology, like hour, color (hue), angle of rotation, biological cycles - all having periodic nature. For example, in the Fig. \ref{fig:faces_circ} there are two natural paths between different hues or rotations, combining them we topologically get toroidal space of features.

We study this problem by proposing a latent space that has enforced a periodic part: considered multiple angles forming a torus, with enforced nearly uniform distribution. Thanks of that, geodesics being paths locally maximizing likelihood, become just segments - are very simple to find. Additionally, thanks to topology of torus, we can choose alternative very different geodesics going through some edge of $[-\pi,\pi]^d$. Hence, we are discussing here this approach from educative perspective: for morphing using alternative paths.



This paper is a proof of concepts for possibility of using latent spaces of more complex topology. Contributions are:
\begin{itemize}[nosep]
	\item latent variables that have periodic nature (Section 2),
  \item spring loss that enforces uniform distribution on periodic latent variables (Section 3),
  \item performing morphing on multiple geodesics paths on torus (Section 4).
\end{itemize}

In future we plan to explore more practical applications for this new possibility. For example for image compression, as, in contrast to vanilla AE used e.g. in WaveOne~\cite{waveone} compressor, uniform distribution on torus allows for very convenient quantization and encoding. Other family of applications is trying to train real-life features with nontrivial topology, for example to be able to automatically recognize rotation of object with circular features, also in 3D if additionally including spherical features.

\section{Construction of the Toroidal AutoEncoder}

We will describe our exemplary implementation of autoencoder on MNIST \cite{mnist} data set. All the code will be made publicly available.

\subsection{Encoder}
The encoder part consists of 3 convolutional layers, each followed by a max-pooling, and one dense layer which we will call encoder output. We construct two identical dense layers of size $d$ called $X$ and $Y$, which are independently plugged to the encoder output. They have no non-linearity, so are able to explore whole $\mathbb{R}^{2d}$. Those $X$ and $Y$ concatenated together make the latent vector of size $2d$. So far, we have just a standard encoder, but with the latent sliced into $X$ and $Y$.

\subsection{Regularization of the latent}
We treat pairs $(x_i,y_i)_{i=1..d}$ of points form $X$ and $Y$ as coordinates on a plane. We do a transition to polar coordinates by:
\be
	r_i^2 =x_i^2+y_i^2\qquad\quad \varphi_i =\textrm{atan2}(y_i,x_i)
\ee
where $\textrm{atan2}$ is 2-argument arctangent with codomain $[-\pi, \pi]$. Denote $R=(r_i)_{i=1..d},\ \Phi=(\varphi_i)_{i=1..d}$.

We regularize 
radii to be from normal distribution $N(1,0.1)$,  
angles to be from uniform distribution on $[-\pi,\pi]$. This regularization is discussed in Section 3.

\subsection{Auxiliary classifier}
A simple classifier consisting of 3 dense layers is connected to $\Phi$. Its sole purpose is to encourage better separation of different classes on d-torus. Its loss (categorical crossentropy) is added to the overall loss. It is not necessary for autoencoder to work, but it shows how we can add some meaning to latent variables. Much more detailed and sophisticated methods are possible for richer data sets, e.g. with pictures of people faces one could make one variable reflect the hair color and make other variable reflect the age of a person on a picture, by attaching multiple auxiliary classifier, each trying to distinguish a single feature.

\subsection{Decoder}
The decoder is roughly symmetric to the encoder, it consists of a dense layer connected to the latent. Output of the dense layer is then reshaped and connected to stack of convolution and up-sampling layers. Note that $\Phi$ and $R$ are not used in decoding - they are used only in order to shape the latent space.

\begin{figure}[h!]
	\centering
	\includegraphics{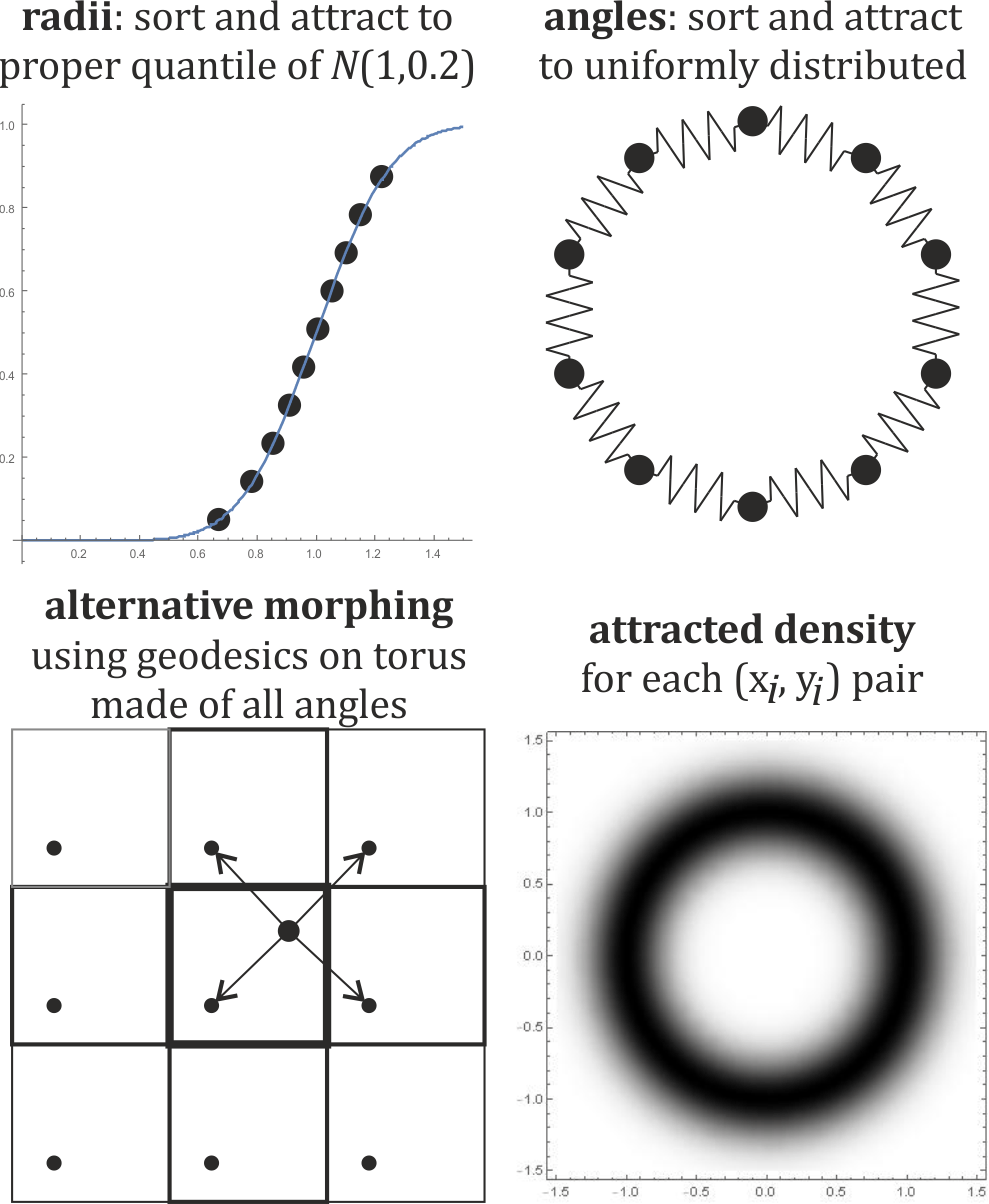}
	\caption{Illustration of spring loss construction}
	\label{quadro}
\end{figure}

\section{Spring loss}

\begin{figure*}[ht]
	\centering
		\includegraphics[width=1.00\textwidth]{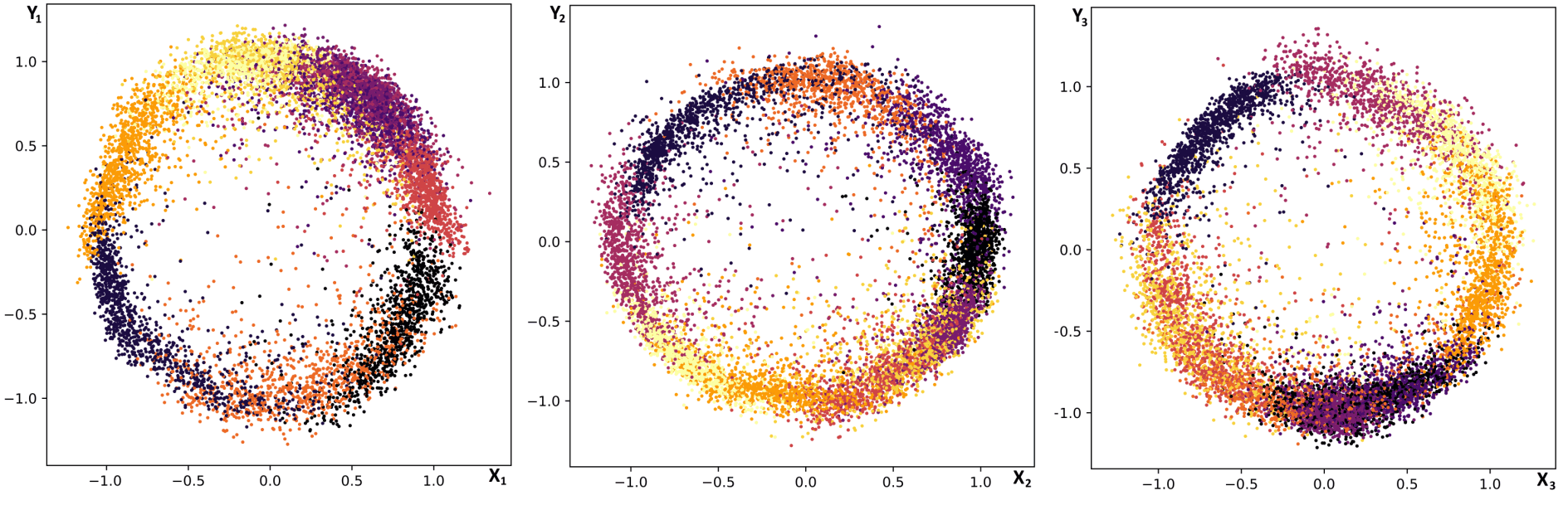}
	\caption{Scatters plots of latent distributions. Colors indicate different classes (digits).}
	\label{fig:3ringsA}
\end{figure*}

We need to introduce additional index which indexes samples in a mini-batch of size $S$. So $\varphi_i^s$ means i-th angle in the latent for s-th sample from the minibatch. We sort all the samples along batch dimension. By $O$ and $o$ we note orders:
	$$r_i^{o_{i,1}}\leq r_i^{o_{i,2}}\leq \ldots \leq r_i^{o_{i,S}}$$
	$$\varphi_i^{O_{i,1}}\leq \varphi_i^{O_{i,2}}\leq \ldots \leq \varphi_i^{O_{i,S}}$$

The spring loss is a sum of squares of distances between sorted values. So intuitively we can imagine a circle made of little balls connected by stretched springs and put around a cylinder  - they all try to attract each other, but the sum of distances is constant. So it turns out that energy is minimal if the balls are equally spaced. Exact formula is the following:
\be
	\mathcal{L}_{spring} = \sum_{i=1}^d\left[(\varphi_i^{O_{i,1}} + \tau -  \varphi_i^{O_{i,S}})^2+\sum_{s=1}^{S-1} \left( \varphi_i^{O_{i,s}}  -  \varphi_i^{O_{i,s+1}} \right)^2\right]
	\label{eq_spring}
\ee
where $\tau = 2 \pi$ is the period \cite{tau}.

Regularization of $R$ is performed by a method very similar to the one proposed in SWAE. Our method is simpler but worse - it does not require sampling from the assumed distribution, but unfortunately does not forbids correlations between variables (we will discuss it in more detail in Section 5). By $q_s$ we note quantiles of wanted distribution. Number of quantiles is equal to the batch size:
\be
	q_s = \textrm{CDF}^{-1}\left(\frac{s-\frac{1}{2}}{S}\right)
\ee
where CDF means cumulative distribution function. For the normal distribution exact formula for the inverse of CDF is:
\be
	\textrm{CDF}^{-1}_{N(\mu,\sigma)}(z) = \mu + \sqrt{2}\ \sigma\ \textrm{erf}^{-1}(2 z -1)
\ee
where $\textrm{erf}^{-1}$ is inverted error function.

We want our radii\footnote{Precisely we use square of the radius in order to avoid instability of derivative of square root in proximity of zero.} to be from normal distribution with $\mu=1$ and $\sigma=0.1$. Minimization of quantile loss defined as:
\be
	\mathcal{L}_{quantile}=\sum_{i=1}^d\sum_{s=1}^S \left(r_i^{o_{i,s}}-q_s\right)^2
\ee
attracts radii to the chosen distribution.

\section{Multiple-path morphing}


\begin{figure*}[tp]
\centering
		\includegraphics[width=1\textwidth]{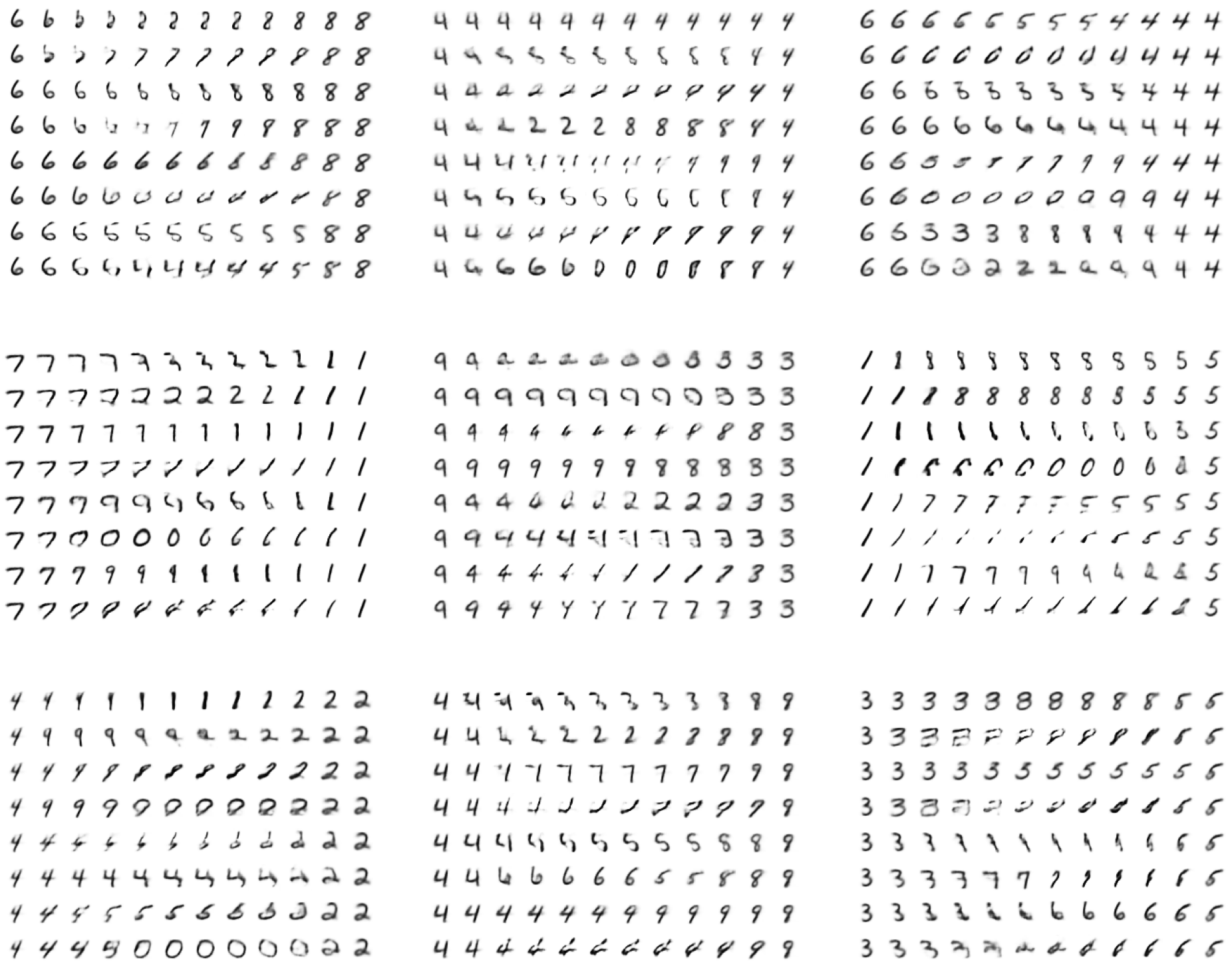}
\caption{Nine examples of multiple-path morphings. Each morphing consists of 8 possible paths (rows) corresponding to different choice of values $P$ from equation (\ref{eq_morph}) in each of three dimensions. Each interpolation path consists of 12 images.}
\label{morph}
\end{figure*}

Interpolation in the latent space is usually performed in this way:
\begin{enumerate}[label=\alph*.,nosep]
	\item two samples $X^{(1)}$ and $X^{(2)}$ are chosen from the validation set,
	\item using encoder they are transformed into latent space to $Z^{(1)}$ and $Z^{(2)}$,
	\item interpolation $Z(t)$ is calculated from formula $Z(t) = (1-t) Z^{(1)} + t\, Z^{(2)}$ for a few equally distributed $t$ in $[0,1]$ range,
	\item these $Z(t)$ are passed through decoder.\\
\end{enumerate}

In our approach points (a) and (b) are the same. Then we transfer latent variables into polar coordinates - denote them by  $\Phi^{(i)},\ R^{(i)}$ for $Z^{(i)}$. For simplicity we do standard linear interpolation (c) for radii, what is approximation of geodesic in their multivariate normal distribution.

However, in the space of angles $\Phi^{(1)},\Phi^{(2)}\in[-\pi,\pi]^d$, we have modulo $2\pi$ arithmetic in each direction. While geodesic in uniform distribution is just a segment, as visualized in Fig. \ref{quadro}, this segment can be inside $[-\pi,\pi]^d$, or can go through one of edges/faces, exploiting cyclic boundary conditions. We could find the closes one: minimizing $\|h_1'-h_2'+2\pi k\|$ for $k\in \mathbb{Z}^d$ and $h'$ being angular coordinates of $h$. However, other segments are also paths locally maximizing likelihood in latent space (assuming uniform distribution).

Finally figure \ref{morph} presents interpolation along a few such segments given by
$$ R(t) = (1-t) R^{(1)} + t\, R^{(2)}$$
\be
	\Phi(t) = (1-t) \Phi^{(1)} + t (\Phi^{(2)}+2k\pi)
	\label{eq_morph}
\ee
for a few different $k\in\{-1,0,1\}^d$, transforming them to Cartesian coordinates for latent variables, and passing through AE decoder.



We performed experiments on MNIST data set using $2d = 6$ dimensional latent space. For each pair of random images from validation set we plotted $2^3$ possible paths. We found it very interesting\footnote{Note that quality of images is not superb, because if relatively small latent space - other experiments usually use latent sizes of 20 or 25.} to observe how each path crosses different classes, see Fig. \ref{morph}. To improve quality we can for example add a few variables enforced to Gaussian distribution.

\section{Discussion and further work}
\begin{figure}[ht]
	\centering
		\includegraphics[width=0.45\textwidth]{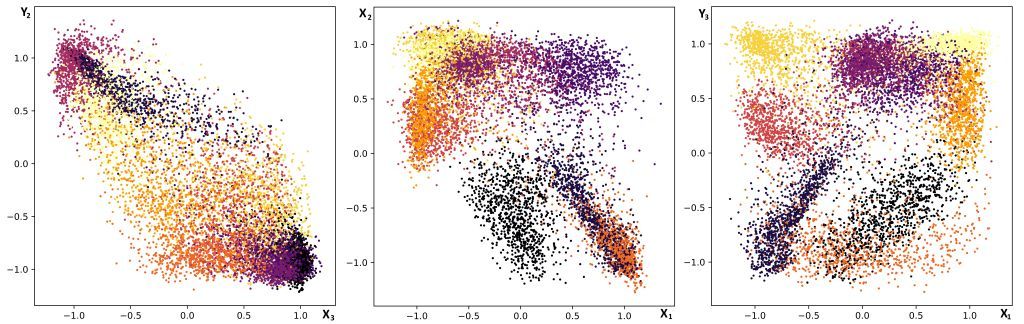}
	\caption{Scatters plots of latent distributions. One can observe pesky correlations and gaps in distributions, corresponding to separation between classes.}
	\label{fig:corr}
\end{figure}

We do admit that current regularization method does not remove correlations, what can be seen in Fig. \ref{fig:corr}. They are unwelcome e.g. due to leaving blank areas in  parts of the latent space. If sampling from such area, or passing morphing through it, the decoder is not able to decode anything decent, because it was not trained on this part of latent space. It is also problematic for planned data compression applications, as uniform distribution would be more convenient to encode.

We plan reduce them for further work for example by using the fact that expected value of product is product of expected values for independent variables, what can be easily transformed to optimization condition for improving independence. Another possibility is using SWAE-like approach: choose random directions and reduce Wasserstein distance between projection and random samples from the desired distribution.

It is worth to mention that recent research by \cite{lesniak} shows that linear interpolation in high dimensional space is not necessarily optimal. Other methods are proposed, that should take into account unintuitive properties of density in high dimensions.

\section{Summary}
This paper addresses subject of periodic latent variables. We argue that some real world features (like time of the day) are naturally periodic and should be represented in a relevant way in the latent space. Motivated by physics, we introduce a new concept of spring loss that enforces uniform distribution with periodic bounds. Finally we show that product of a few periodic latent variables may be interpreted as a torus. This allows alternative geodesics interpolations between points in the latent space.


{\small
\bibliographystyle{ieee}
\bibliography{tae}
}
\end{document}